\documentclass[11pt,a4paper]{article}
\usepackage{times,latexsym}
\usepackage{url}
\usepackage[T1]{fontenc}
\usepackage{xspace}
\usepackage{color}
\usepackage{graphicx}
\usepackage{amsmath}
\usepackage{float}
\usepackage{multirow}
\usepackage{fixltx2e}
\usepackage{booktabs}
\usepackage[table]{xcolor}
\usepackage{xcolor}
\usepackage{booktabs}
\usepackage{soul}

\usepackage[acceptedWithA]{tacl2021v1}

\usepackage{xspace,mfirstuc,tabulary}
\usepackage{enumitem}

\newcommand{\stitle}[1]{\vspace{1ex} \noindent{\bf #1}}

\definecolor{yaleblue}{rgb}{0.06, 0.3, 0.57}

\newcommand{\revision}[1]{{#1}}
\newcommand{\secondrevision}[1]{{#1}}
\newcommand{\modelnamets}{\texttt{LITE}}
\newcommand{\modelname}{\texttt{LITE}\xspace}

\newcommand{\litebeer}{\includegraphics[height=1em]{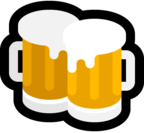}}

\newif\iftaclinstructions
\taclinstructionsfalse % AUTHORS: do NOT set this to true
\iftaclinstructions
\renewcommand{\confidential}{}
\renewcommand{\anonsubtext}{(No author info supplied here, for consistency with
TACL-submission anonymization requirements)}
\newcommand{\instr}
\fi

\iftaclpubformat % this "if" is set by the choice of options
\newcommand{\taclpaper}{final version\xspace}

\else
\newcommand{\taclpaper}{submission\xspace}

\fi

\title{Ultra-fine Entity Typing with Indirect Supervision from Natural Language Inference}

\author{
   Bangzheng Li$^{\diamond\dagger}$\Thanks{This work was done when the first author was visiting
the University of Southern California.}$\;$, Wenpeng Yin$^\ddag$ \and Muhao Chen$^\diamond$ 
   \\
   $^\diamond$University of Southern California\\
   $^\dagger$University of Illinois at Urbana-Champaign\\
   $^\ddag$Temple University\\
   \texttt{vincentleebang@gmail.com}; $\;$\texttt{wenpeng.yin@temple.edu};\\
   \texttt{muhaoche@usc.edu}
 }

\date{Feb 9, 2021}

\usepackage{cleveref}
\crefformat{section}{\S#2#1#3}
\crefformat{subsection}{\S#2#1#3}
\crefformat{subsubsection}{\S#2#1#3}
\crefrangeformat{section}{\S\S#3#1#4 to~#5#2#6}
\crefmultiformat{section}{\S\S#2#1#3}{ and~#2#1#3}{, #2#1#3}{ and~#2#1#3}
\usepackage{refstyle}
\Crefformat{figure}{#2Fig.~#1#3}
\Crefmultiformat{figure}{Figs.~#2#1#3}{ and~#2#1#3}{, #2#1#3}{ and~#2#1#3}
\Crefformat{table}{#2Tab.~#1#3}
\Crefmultiformat{table}{Tabs.~#2#1#3}{ and~#2#1#3}{, #2#1#3}{ and~#2#1#3}
\Crefformat{appendix}{Appx.~\S#2#1#3}
\crefformat{algorithm}{Alg.~#2#1#3}

\begin{document}
\maketitle
\begin{abstract}
  The task of ultra-fine entity typing (UFET) seeks to predict diverse and free-form words or phrases that describe the appropriate types of entities mentioned in sentences. A key challenge for this task lies in the large amount of types and the scarcity of annotated data per type. Existing systems formulate the task as a multi-way classification problem and train directly or distantly supervised classifiers. This causes two issues: 
  (i) the classifiers do not capture the type semantics since types are often converted into indices; (ii) systems developed in this way are limited to predicting within a pre-defined type set, and often fall short of generalizing to types that are rarely seen or unseen in training.

  This work presents \modelname~\litebeer, a new approach that formulates entity typing as a natural language inference (NLI) problem, making use of (i) the indirect supervision from NLI to infer type information meaningfully represented as textual hypotheses and alleviate the data scarcity issue, as well as (ii) a learning-to-rank objective to avoid the pre-defining of a type set. Experiments show that, with limited training data, \modelname obtains state-of-the-art performance on the UFET task. In addition, \modelname demonstrates its strong generalizability, by not only yielding best results on other fine-grained entity typing benchmarks, more importantly, a  pre-trained \modelname system works well on new data containing unseen types.\footnote{Our models and implementation are available at \url{https://github.com/luka-group/lite}.}
\end{abstract}

\section{Introduction}
% \color{blue}[test]
% 1. Background: what is entity typing? why addressing this problem?
% 2. Why ultra-fine entity typing? Why challenging? Why existing works fall short?
% 2.1 Very large candidate spaces
% 2.2 Insufficient training labels
% 2.3 Additional challenges: such as dependency of different levels of labels ...

Entity typing, inferring the semantic types of the entity mentions in text, is a fundamental and long-lasting research problem in natural language understanding, which aims at inferring the semantic types of the entities mentioned in text. The resulted type information can help with grounding human language components to real-world concepts \cite{chandu2021grounding},
and provide valuable prior knowledge for natural language understanding tasks such as entity linking~\cite{ling-etal-2015-design,onoe-durrett-2020-interpretable}, question answering~\cite{yavuz-etal-2016-improving}, and information extraction~\cite{koch-etal-2014-type}. Prior studies have mainly formulated the task as a multi-way classification problems~\cite{Wang2021KAdapterIK,zhang-etal-2019-ernie,Chen2020HierarchicalET,9305269}.  
% The token-level classifiers developed in those studies have performed well on less fine-grained entity typing benchmarks, including FIGER~\cite{ling-etal-2015-design} and Ontonotes~\cite{gillick2014context}, where the number of types range from dozens to hundreds. 

\begin{figure*}[t!]
\begin{center}
    \includegraphics[scale=0.38]{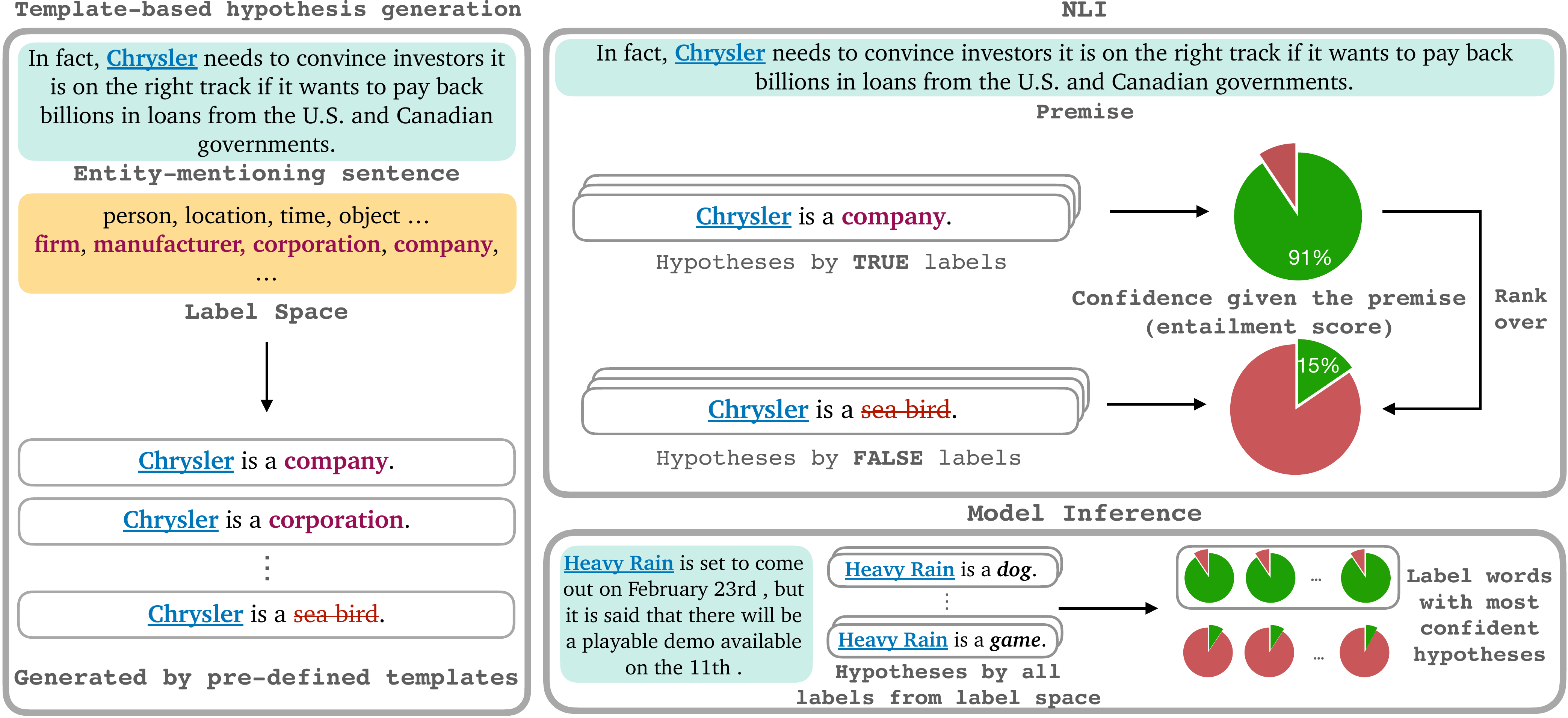}
    %\vspace{-2em}
    \caption{Entity typing by \modelname with indirect supervision from NLI.}
    \label{fig:architecture}
    %\vspace{-1em}
\end{center}
\end{figure*}

However, earlier efforts for entity typing are far from enough for representing real-world scenarios, where types of entities can be extremely diverse.
Accordingly, the community has recently paid much attention to more fine-grained modeling of types for entities. One representative work is the Ultra-fine Entity Typing (UFET) benchmark created by \citet{choi-etal-2018-ultra}. The task seeks to search for the most appropriate types for an entity among over ten thousand free-form type candidates. The drastic increase of types enforces us to doubt if the multi-way classification framework is still suitable for UFET. In this context, two main issues are noticed from prior work. First, prior studies have not tried to understand the target types since most classification systems converted all types into indices. 
% Without knowing the semantics of types, it is unlikely to match an entity mention to a correct type especially when the type does not have rich annotations.
\revision{Without knowing the semantics of types, it is hard to match an entity mention to a correct type especially when there is not %extra information about the type words
sufficient annotated data for each type.} 
Second, existing entity typing systems are far behind the desired capability in real-world applications in which any open-form types can appear. 
% Unfortunately, those pre-trained multi-way classifiers cannot recognize types that are unseen in training. For example, a system trained on UFET even needs retraining if we want to deploy it to a coarser-grained entity typing task, e.g., Ontonotes \cite{gillick2014context} and FIGER \cite{ling2012fine}.
\revision{Specifically, those pre-trained multi-way classifiers cannot recognize types that are unseen in training, especially when there is no reasonable mapping from existing types to unseen type labels, %In this case, re-training with enlarged type vocabulary is necessary for them to perform well. 
unless the classifiers are re-trained to include those new types.
}

% which has obtained high-quality human annotations with a label space of over ten thousand free-form entity type labels. 
% As a result, this task features the inference of much more diverse and concrete type information for entity mentions than traditional entity typing tasks. At the same time, this task introduces a non-trivial learning problem that causes previous entity typing models to easily fall short.
% This is not only due to much higher cost for a model to perform precise inference on an extremely large candidate space.
% More importantly, the tremendous type vocabulary also leads to insufficient training data per entity type, due to the cost of obtaining high-quality annotations.
% Hence, this resource-hungry task also challenges learning-based entity typing model with a low-resource and extremely few-shot learning scenario \cite{raykar2010learning}.
%it is costly to obtain high-quality annotated examples for each type. In other words, this task is resource-hungry which prevents the previous works from performing well.

% 3. We propose a new (different solution). What kind of solution? What benefits does it have?
% 4. What is the outcome of this work.
% 5. A list of contributions... (At least 3 Orthogonal.)

To alleviate the aforementioned challenges, we propose a new learning framework that seeks to enhance ultra-fine entity typing with indirect supervision from natural language inference (NLI) \cite{10.1007/11736790_9}. Specifically, our method \modelname~\litebeer, namely (\underline{L}anguage \underline{I}nference based \underline{T}yping of \underline{E}ntities), treats each entity-mentioning sentence as a premise in NLI. 
Using simple, template-based generation techniques, a candidate type is transformed into a textual description and is treated as the hypothesis in NLI.
Based on the premise sentence and a hypothesis description of a candidate type, the entailment score given by an NLI model is regarded as the confidence of the type. On top of the pre-trained NLI model, \modelname conducts a learning-to-rank objective, which aims at scoring hypotheses of positive types higher than the hypotheses of sampled negative types. 
Finally, the label candidates whose hypotheses obtain scores above a threshold are given as predictions by the model.

Technically, \modelname benefits ultra-fine entity typing from three perspectives.
First, the inference ability of a pre-trained NLI model can provide effective indirect supervision to improve the prediction of type information. 
Second the hypothesis, as a type description, also provides a semantically rich representation of the type, which further benefits few-shot learning with insufficient labeled data.
Moreover, to handle the dependency of type labels in different granularities, we also utilize the inference ability of NLI model to learn that the finer label hypothesis of an entity mention entails its general label hypothesis. 
Experimental results on the UFET benchmark~\cite{choi-etal-2018-ultra} show that \modelname drastically outperforms the recent state-of-the-art (SOTA) systems~\cite{Dai2021UltraFineET,Onoe2021ModelingFE,liu2021finegrained} without any need of distantly supervised data as they do. 
% by improving \modelname of F-1 score. 
%In addition, we also apply %our training method to traditional (less) fine-grained entity typing tasks, including FIGER~\cite{ling-etal-2015-design} and Ontonotes~\cite{gillick2014context}, where it also achieves best performances. 
In addition, our \modelname also yields the best performance on traditional (less) fine-grained entity typing tasks.\footnote{Note that although these more traditional entity typing tasks are termed as ``fine-grained entity typing'', their typing systems are much less fine-grained than that of UFET.}
% What’s more, to evaluate our approach in a more open-form scenario, we truncate the labeling vocabulary and our model still achieves highest performances on the new data.
What's more, since we adopt a learning-to-rank objective to optimize the inference ability of \modelname rather than classification on a specified label space, it is feasible to apply the trained model across different typing data sets. We therefore test its transferability by training on UFET and evaluate on traditional fine-grained benchmarks to get promising results. 
\secondrevision{Moreover, we also examined the time efficiency of \modelname, and discussed about the trade-off between training and inference costs in comparison with prior methods.}

To summarize, the contributions of our work are three-folds.
First, to our knowledge, this is the first work that uses NLI formulation and NLI supervision to handle entity typing. 
As a result, our system is able to keep the labels' semantics and encode the label dependency effectively. 
Second, our system offers SOTA performance on both ultra-fine entity typing and regular fine-grained typing tasks, being particularly strong at predicting zero-shot and few-shot cases. 
Finally, we show that our system, once trained, can also work on  different test sets which are free to have unseen types.

\section{Related Work}
% Our work is connected to two %major topics in NLP.
% research topics.
% Each has a large body of work which we can only provide as a highly selected summary.

\stitle{Entity Typing.} 
Traditional entity typing was introduced and thoroughly studied by \citet{ling2012fine}. 
One main challenge that earlier efforts have focused on was to obtain sufficient training data to develop the typing model. To do so, automatic annotation has been commonly used in the a series of works~\cite{gillick2014context,ling2012fine,yogatama-etal-2015-embedding}. 
Later works were developed for further improvement by modeling the label dependency with a hierarchy-aware loss~\cite{Ren2016LabelNR, Xu2018NeuralFE}. 
External knowledge from knowledge bases has also been introduced to capture the semantic relations or relatedness of type information~\cite{jin-etal-2019-fine,dai-etal-2019-improving,obeidat-etal-2019-description}. 
\citet{ding2021promptlearning} adopt prompts to model the relationship between entities and type labels, which is similar to our template-based type description generation. However, their prompts are intended for label generation from masked language models while our templates realize the supervision from NLI.

More recently, \citet{choi-etal-2018-ultra} proposed the ultra-fine entity typing (UFET) task which involved free-form type labeling to realize the open-domain label space with much more comprehensive coverage of types.
As the UFET tasks non-trivial learning and inference problems, several methods have been explored by more effectively modeling the structure of the label space. \citet{xiong-etal-2019-imposing} utilized a graph propagation layer to impose label-relation bias in order to capture type dependencies implicitly. \citet{onoe-durrett-2019-learning} trained a filtering and relabeling model with the human annotated data to denoise the automatically generated data for training.
\citet{Onoe2021ModelingFE} introduced box embeddings \cite{vilnis-etal-2018-probabilistic} to represent the dependency among multiple levels of type labels as topology of axis-aligned hyper-rectangles (\emph{boxes}). 
To further cope with insufficient training data,
\citet{Dai2021UltraFineET} used pre-trained language model for augmenting (noisy) training data with masked entity generation. 
% However, the existing works mostly treat this problem as a traditional classification task while our approach formulate it in a ranking manner.
Different to their strategy of augmenting training data, our approach generates type descriptions to leverage indirect supervision from NLI which requires no more data samples.

\stitle{Natural Language Inference and Its Applications.}
% \wenpeng{This part should be sometihng like "Natural language inference for downstream NLP tasks" instead of "Natural language inference" since we are not trying to solve NLI itself.}
Early approaches towards NLI problems were based on studying lexical semantics and syntactic relations~\cite{10.1007/11736790_9}. 
Following research then introduced deep-learning methods into this task to capture contextual semantics. \citet{parikh-etal-2016-decomposable} utilize Bi-LSTM~\cite{hochreiter1997long} to encode the input tokens and use attention mechanism to capture substructures of input sentences. 
Most recent works develop end-to-end trained NLI models that leverage pre-trained language models \cite{devlin-etal-2019-bert, liu2019roberta} for sentence pair representation and large learning resources \cite{bowman-etal-2015-large, williams-etal-2018-broad} for training. 

Specifically, since pre-trained NLI models benefit generalizable logical inference, current literature has also proposed to leverage NLI models to improve prediction tasks with insufficient training labels, including zero-shot and few-shot text classification \cite{yin-etal-2019-benchmarking}. 
\citet{Shen2021TaxoClassHM} adopted \emph{RoBERTa-large-MNLI} \cite{liu2019roberta} to calculate the document similarity for document multi-class classification. 
\revision{\citet{chen2021nli} proposed to verify the output of a QA system with NLI models by converting the question and answer into a hypothesis and extracting textual evidence from the reference document as the premise.}

% A recent work by \citet{yin-etal-2020-universal} is particularly relevant to this topic, which utilize NLI as a unified solver for several text classification tasks such as co-reference resolution and multiple choice QA in a few-shot manner. 
\revision{Recent works by \citet{yin-etal-2020-universal} and \citet{white-etal-2017-inference} are particularly relevant to this topic, which utilize NLI as a unified solver for several text classification tasks such as co-reference resolution and multiple choice QA in few-shot or fully-supervised manner.} 
Yet our work handles a learning-to-rank objective for inference in a large candidate space, which not only enhances learning under a data-hungry condition, but also is free to be adapted to infer new labels that are unseen to training.  
\revision{\citet{yin-etal-2020-universal} also proposed an approach to transform co-reference resolution task into NLI manner and we modified it as one of our template generation methods, which is discussed in \Cref{ssec:temp}}.

% \usepackage{booktabs}
\begin{table*}[t]
\scriptsize
\setlength\columnsep{0pt}
\centering
\begin{tabular}{p{0.17\linewidth}|p{0.23\linewidth}|p{0.52\linewidth}}
\toprule
\multicolumn{1}{c|}{Templates} & \multicolumn{1}{c|}{Type Descriptions}                         & \multicolumn{1}{c}{Premise-Hypothesis Pairs for NLI}                                                                                                                                    \\ \midrule \midrule
Taxonomic Statement            &  \underline{\textbf{Jay}} is a \textbf{producer}.      & \begin{tabular}[c]{@{}l@{}}Premise: ``\underline{\textbf{Jay}} is currently working on his Spring 09 collection, …''\\ Hypothesis: ``\underline{\textbf{Jay}} is a \textbf{producer}.''\end{tabular}                                       \\ \midrule
Contextual Explanation         & In this context, \underline{\textbf{career at a com-}} \underline{\textbf{-pany}} is referring to \textbf{duration}. & \begin{tabular}[c]{@{}l@{}}Premise:  ``No one expects a \underline{\textbf{career at a company}} any more, …''\\ Hypothesis: ``In this context, \underline{\textbf{career at a company}} is referring to \textbf{duration}.''\end{tabular} \\ \midrule 
Label Substitution             & \textbf{Musician} knows how to make a hip-hop record sound good.        & \begin{tabular}[c]{@{}l@{}}Premise: ``\underline{\textbf{He}} knows how to make a hip-hop record sound good.''\\ Hypothesis: ``\textbf{Musician} knows how to make a hip-hop record sound good.'' 
\end{tabular}       \\ \bottomrule
\end{tabular}
%\vspace{-0.5em}
\caption{Type description instances of three templates. \underline{\textbf{Entity mentions}} are boldfaced and underlined while \textbf{label words} are only boldfaced.}
\label{tab:typeDes}
%\vspace{-1em}
\end{table*}
% \begin{figure*}[t]
% \begin{center}
%     \includegraphics[height=150pt, width=480pt]{figures/typeDescription.pdf}
% \end{center}
% \caption{\label{tab:temp} template}
% \end{figure*} 

\section{Method}
% \muhao{A paragraph is needed at here to describe what is introduced in this section.}

In this section, we introduce the proposed method for (ultra-fine) entity typing with NLI. We start with the preliminary of problem definition and the overview of our NLI-based entity typing framework (\Cref{ssec:probDef}), followed by technical details of type description  generation  (\Cref{ssec:temp}), label dependency modeling (\Cref{ssec:depend}), learning objective (\Cref{ssec:obj}) and inference (\Cref{ssec:inf}).

\subsection{Preliminaries}\label{ssec:probDef}

\stitle{Problem Definition.}
The input of an entity typing task is a sentence $s$ and an entity mention of interest $e\in s$. This task aims at typing $e$ with one or more type labels from the label space $L$. For instance, in \textit{``\underline{Jay} is currently working on his Spring 09 collection , which is being sponsored by the YKK Group.’’}, the entity ``\textit{Jay}'' should be labeled as \textit{person}, \textit{designer} or \textit{creator} instead of \textit{organization} or \textit{location}.

The structure of the label space $L$ can vary. For example, in some benchmarks like OntoNotes \cite{gillick2014context}, labels are provided in canonical form and strictly depend on their ancestor types. In this case, a type label \emph{bridge} appears as \emph{/location/transit/bridge}. However, in benchmarks like FIGER \cite{ling2012fine}, partial labels have a dependency with their ancestors while the others are free-form and uncategorized. For instance, label \emph{film} is given as \emph{/art/film} but \emph{currency} appears as a single word. For our primary task, for ultra fine-grained entity typing, the UFET benchmark ~\cite{choi-etal-2018-ultra} provides no ontology of the labels and the label vocabulary consists of free-form words only. In this case, \emph{film star} and \emph{person} can appear independently in an annotation set with no dependency information provided.

\stitle{Overview of \modelname.}
Given a sentence with at least an entity mention, \modelname treats the sentence as the premise in NLI, and then learns to type the entity in three consecutive steps (\Cref{fig:architecture}).
First, \modelname employs a simple, low-cost template-based technique to generate a natural language description for a type candidate. 
This type description is treated as the hypothesis in NLI.
For this step, we explore with three different description generation templates (\Cref{ssec:temp}).
Second, to capture label dependency, whether or not the type ontology is provided, \modelname consistently generates type descriptions for any ancestors of the original type label on the previous sentence and learns their logical dependencies (\Cref{ssec:depend}).
These two steps create positive cases of type descriptions for the entity mention in the previous sentence.
Last, \modelname fine-tunes a pre-trained NLI model with a learning-to-rank objective that ranks the positive case(s) over negative-sampled type descriptions according to the entailment score (\Cref{ssec:inf}).
During the inference phase, given another sentence that mentions an entity to be typed, our model predicts type that leads to the hypothetical type description with the highest entailment score.
In this way, \modelname can effectively leverage indirect supervision signals of a (pre-trained) NLI model to infer the type information of a mentioned entity.

We hereby describe the technical details of training and inference steps of \modelname in the rest of the section.

\subsection{Type Description Generation}\label{ssec:temp} 
Given each sentence $s$ with an annotated entity mention $e$, \modelname first generates a natural language type description $T(a)$ for the type label annotation $a$.  The description will later act as a hypothesis in NLI. 
Specifically, we consider several generation technique to obtain such type descriptions, for which the details are described as follows.

\begin{itemize}[leftmargin=1em]
    \item \emph{Taxonomic statement.} The first template directly connects the entity mention and the type label with an ``is-a'' statement, i..e. ``\texttt{[ENTITY]} is a \texttt{[LABEL]}''. 
    \item \emph{Contextual explanation}. The second template generates a declarative sentence which adds a context-related connective. The generated type description is in the form of ``In this context, \texttt{[ENTITY]} is referring to \texttt{[LABEL]}''.
    \item \emph{Label substitution}. \revision{\citet{yin-etal-2020-universal} proposed to transform co-reference resolution problem into NLI manner by replacing the pronoun mentions with candidate entities. Inspired by their transformation, this technique} directly replaces the \texttt{[ENTITY]} in the original sentence with \texttt{[LABEL]}. Therefore, the NLI model will treat the modified sentence with a ``type mention'' as the hypothesis of the original sentence with the entity mention. 
    
\end{itemize}

\noindent
As shown in \Cref{tab:typeDes}, each template provides a semantically meaningful way to connect the entity with label. In this way, the inference ability of an NLI model can be leveraged to capture the relationship of entity and label, given the original entity-mentioning sentence as the premise. 

%To obtain effective templates for \modelname, we have also tried 
Particularly, we have also tried automatic template generation method proposed by \citet{gao2021making}, which has led to the adoption of the contextual explanation template. Such a template technique adopts the pre-trained text-to-text Transformer T5 \cite{2020t5}  to generate prompt sentences for fine-tuning language models. In our case, T5 mask tokens are added between the sentence, the entity and the label. Since T5 is trained to fill in the blanks within its input, the output tokens can be used as the template for our type description. 
For example, given the sentence \textit{``Anyway, Nell is their new singer, and I would never interrupt her show.''}, the entity \textit{Nell} and the annotations \textit{(singer, musician, person)}, we can formulate the input to T5 as \textit{``Anyway, Nell is their new singer, and I would never interrupt her show. <X> Nell <Y> singer <Z>''}. T5 will then fill in the placeholders \textit{<X>, <Y>, <Z>} and output ``... I would never interrupt her show. In fact, Nell is a singer.'' 
We observe that most of the generated templates given by T5 have appeared as the format where a prepositional phrase (e.g. \emph{in fact}, \emph{in this context}, \emph{in addition}, etc.) followed by a statement such as \textit{``[ENTITY] is a [LABEL]''} or \textit{``[ENTITY] became [LABEL]''}. 
Accordingly, we select the above contextual explanation template, which is the most representative pattern observed in the generations.

\iffalse
When decoding, we aim to find templates that are suitable for all data samples in the train set. Formally, given an input sample $(s,e,a)$ from the train set $D_{train}$, we adopt the objective proposed by \cite{gao2021making} finding out the template $T$ that maximizes \[\sum_{(s,e,a)\in D_{train}}\log P_{T5}(T|T(s,e,a))\] where $P_{T5}$ denotes the output probability distribution of T5. Accordingly, it can be decomposed to: 
\[
    \sum^{|T|}_{j=1}\sum_{(s,e,a)\in D_{train}}\log P_{T5}(t_j|t_1,..., t_{j-1}, T(s,e,a))
\] 
where $|T|$ is a predefined number of total generated tokens and $(t_1,..., t_j)$ are the template tokens. 

We use beam search of width 50 to obtain diverse template candidates. The corresponding results are similar in the sense of format: a prepositional phrase (e.g. In fact, In addition, In the end...) or a connective (e.g. But, However...) followed by a statement such as \textit{``[ENTITY] is a [LABEL]''} or \textit{``[ENTITY] became [LABEL]''}. Since it is expensive to test all of these similar templates, we select the \emph{contextual explanation} template as one of the three main templates. 
\fi

% In our experiment, we have observed that, among the three description generation templates, the taxonomic statement generation generally gives better performance than the others under most settings, for which the analysis is presented in \Cref{sec:analysis}. Thus, the main experimentation is reported as the configuration where \modelname uses the type descriptions based on taxonomic statement. 
\revision{In the training process, we use one of the three templates to generate the hypotheses, for which the same template will also be used to obtain the candidate hypotheses in inference. 
According to our preliminary results on dev set,} the taxonomic statement generation generally gives better performance than the others under most settings, for which the analysis is presented in \Cref{sec:analysis}. Thus, the main experimentation is reported as the configuration where \modelname uses the type descriptions based on taxonomic statement. 

\subsection{Modeling Label Dependency}\label{ssec:depend}
The rich entity type vocabulary may form hierarchies that enforce logical dependency among labels of different specificity. Hence, we extend the generation process of type description to better capture such the label dependency. 
In detail, for a specific type label that \modelname has generated a type description, if there are ancestor types, we not only generate descriptions for each of the ancestor types, but also conduct learning among these type descriptions. The descendant type description would act as the premise and the ancestor type description would act as the hypothesis. For instance, in OntoNotes \cite{gillick2014context} or FIGER \cite{ling2012fine}, suppose a sentence mentions the entity \emph{London} and is labeled as \emph{/location/city}, if the taxonomic statement based description generation is used, \modelname will yield descriptions for both levels of types, i.e. ``London is a city'' and ``London is a location''. 
\revision{In such a case, the more fine-grained type description ``London is a city'' can act as the premise of the more coarse-grained description ``London is a location'', so as to help capturing the dependency between two labels ``city'' and ``location''.} Such paired type descriptions are added to training and will be captured by the dependency loss $\mathcal{L}_{d}$ as being described in \Cref{ssec:obj}. 

This technique to capture label dependency can be easily adapted to tasks where a type ontology is unavailable, but each instance is directly annotated with multiple type labels of different specificity. Particularly for the UFET task~\cite{choi-etal-2018-ultra}, while no ontology is provided for the label space, the task separates the type label vocabulary into different specificity, i.e. \emph{general}, \emph{fine} and \emph{ultra-fine} ones. Since its annotation to an entity from a sentence includes multiple labels of different specificity, we can still utilize the aforementioned dependency modeling method. For instance, an entity \emph{Mike Tyson} may be simultaneously labeled as \emph{person} (general), \emph{sportsman} (fine), and \emph{boxer} (ultra-fine). 
Similar to using an ontology, each pair of descendant and ancestor descriptions among the three generations ``Mike Tyson is a sportsman'', ``Mike Tyson is a person'' and ``Mike Tyson is a sportsman'' are also added to training.

\subsection{Learning Objective}\label{ssec:obj}
Let $L$ be the type vocabularies, the learning objective of \modelname is to conduct learning-to-rank on top of the NLI model.
Given a sentence $s$ with mentioned entity $e$,
we use $P$ to denote all true type labels of $e$ that may include the original label and any induced ancestor labels as described in \Cref{ssec:depend}.
Then, for each label $p\in P$ whose type description is generated as $H(p)$ by one of the techniques in \Cref{ssec:temp},
the NLI model calculates the entailment score $\varepsilon (s,H(p)) \in [0,1]$ for the premise $s$ and hypothesis $H(p)$. 
Meanwhile, negative sampling randomly selects a false label $p^\prime\in L\setminus{P}$. Following the same procedure above, the entailment score $\varepsilon (s,H(p^\prime))$ is obtained for the premise $s$ and the negative-sample hypothesis $H(p^\prime)$. The margin ranking loss for an annotated training case is then defined as
\begin{align*}
    \mathcal{L}_{t}=[ \varepsilon(s,H(p^\prime))-\varepsilon(s,H(p))+\gamma ]_+.
\end{align*}
$[x]_{+}$ denotes the positive part of the input $x$ (i.e. $max(x,0)$) and $\gamma$ is a non-negative constant. 

We also similarly define a ranking loss to
model the label dependency. Still given the above annotated sentence $s$ and the set of all true type labels $P$, as described in \Cref{ssec:depend}, for any exiting pair of ancestor type $p_{an}$ and descendant type $p_{de}$ from $P$, the training phase also captures the entailment relation between their descriptions. 
This process regards $H(p_{de})$ as the premise and $H(p_{an})$ as the hypothesis, and the NLI model therefore yields an entailment score $\varepsilon (H(p_{de}), H(p_{an}))$.
The label dependency loss is then defined as the following ranking loss
{
\small
\begin{align*}
    \mathcal{L}_{d}=[ \varepsilon (H(p_{de}), H(p_{an}^\prime))-\varepsilon (H(p_{de}), H(p_{an}))+\gamma ]_+,
\end{align*}
}
where $p_{an}^\prime$ is negative-sampled type label.

\iffalse
To put together, for each $a\in A$, we have the learning objective 

\begin{align*}
    \mathcal{L}_{a}=\mathcal{L}_{t}+\lambda\mathcal{L}_{d}.
\end{align*}
where $\lambda$ denotes the hyper-parameter controlling the influence of dependency modeling.
\fi

The eventual learning objective is to optimize the following joint loss:
\begin{align*}
    \mathcal{L}=\dfrac{1}{|S|}\sum_{s\in S}\dfrac{1}{|P_s|}\sum_{p\in P_s}\mathcal{L}_{t}+\lambda\mathcal{L}_{d}
\end{align*}
where $S$ denotes the dataset containing sentences with typed entities, and $P_s$ denotes the set of true labels on an entity of the sentence instance $s$. \revision{In this way, all annotations of each entity mention will be involved in training.}  $\lambda$ here is a non-negative hyper-parameter that controls the influence of dependency modeling.

\subsection{Inference}\label{ssec:inf}
% The inference phase of \modelname performs ranking on descriptions for all candidate type labels. 
\revision{The inference phase of \modelname performs ranking on descriptions for all type labels from the vocabulary.} 
For any given sentence $s$ mentioning an entity $e$, \modelname accordingly generates a type description for each candidate type label. Then, taking the sentence $s$ as the premise, the fine-tuned NLI model ranks the hypothetical type descriptions according to their entailment scores.
Finally, \modelname selects the type label whose description receives the highest entailment score, or predicts with a threshold of entailment scores in cases where multi-label prediction is required.

\section{Experiment}
\label{sec:experiments}

In this section, we present the experimental evaluation for \modelname framework, based on both UFET (\Cref{ssec:ultra}) and traditional (less) fine-grained entity typing tasks (\Cref{ssec:fine}). In addition, we also conduct comprehensive ablation studies to understand the effectiveness of the incorporated techniques(\Cref{sec:analysis}).

\subsection{Ultra-Fine Entity Typing}\label{ssec:ultra}
We use the UFET benchmark created by \citet{choi-etal-2018-ultra} for evaluation. The UFET dataset consists of two parts. (i) Human-labeled data (L): 5,994 instances split into train/dev/test by 1:1:1 (1,998 for each); (ii) Distant supervision data (D): including 5.2M instances that are automatically labeled by linking entity to KB, and 20M instances generated by headword extraction. We follow the original design of the benchmark to evaluate loose macro-averaged precision (P), recall (R) and F1. 

% Its whole labeling space contains 10,331 types within which there are 9 general types, 121 fine-grained types and 10,201 ultra-fine types. The full dataset consists of 5,994 human annotated data, \muhao{+splits}.
% It also comes with 5.2M data automatically labeled by linking entity to KB as well as 20M data generated by headword extraction. Such automatically generated data serve as the distant supervision for training. 

% \subsubsection{Experimental Settings} 

\stitle{Training Data.} In our approach, the supervision can come from the M\textbf{NLI} data (NLI)~\cite{williams-etal-2018-broad}, \textbf{d}istant supervision data (D) and the human-\textbf{l}abeled data (L). Therefore, we investigate the best combination of training data by exploring the following different training pipelines:

% In prior studies, training data for this benchmark, as mentioned before, involves sources of distant supervision data and human annotated data \muhao{+citations}. We accordingly adopt four variants of \modelname to compare with baseline methods that follow different training settings involving such augmented training data: 

\begin{itemize}[leftmargin=1em]
\setlength\itemsep{0em}
    \item \modelnamets\textsubscript{NLI}: Pre-train on MNLI\footnote{This is obtained from \url{huggingface.co/roberta-large-mnli}}, then predict directly, without any tuning on D or L;
    \item \modelnamets\textsubscript{L}: \revision{Only fine-tune on L};
    \item \modelnamets\textsubscript{NLI+L}: Pre-train on MNLI, then fine-tune on L;
    \item \modelnamets\textsubscript{D+L}: Pre-train on D, then fine-tune on L;
    \item \modelnamets\textsubscript{NLI+D+L}: First pre-train on MNLI, then on D, finally fine-tune on L.
    % \item \modelnamets\textsubscript{NLI/D$^+$/L}: It follows the training pipeline of \modelnamets\textsubscript{NLI/D/L} except that we extend D with extra distant supervision data generated by BERT masked language model \cite{Dai2021UltraFineET}. The enlarged distant supervision data is denoted as ``D$^+$''.
    
    % \wenpeng{Since distant supervision does not help for your system, it does not make sense to report an extended distant supervision data performance}
\end{itemize} 

\stitle{Model Configurations.} Our system is first initialized as  RoBERTa-large~\cite{liu2019roberta} and AdamW  \cite{loshchilov2018fixing} is used to optimize the model. The hyper-parameters as well as the output threshold are tuned on the dev set: batch size 16, pre-training (D) learning rate 1e-6,  fine-tuning (L) learning rate  5e-6, margin $\gamma$=0.1 and $\lambda$=0.05. 
% , are chosen from 0 to 0.4 with a step of 0.1. The hyper-parameter $\lambda$ for controlling the influence of dependency modeling are chosen from 0 to 0.5 with a step of 0.05. Accordingly,  $\gamma$=0.1 and $\lambda$=0.05 are selected for the reported \modelname performances. 
The pre-training epochs are limited to 5 that are enough considering the large size of pre-training data. The fine-tuning epochs are limited to 2,000; models are evaluated every 30 epochs on dev and the best model is kept to conduct inference on test. 

\stitle{Baselines.} We compare \modelname with the following strong baselines. Except for LRN which is merely trained on the human annotated data, all the other baselines incorporate the distant supervision data as extra training resource. 

\begin{itemize}[leftmargin=1em]
    \item \textbf{UFET-biLSTM} \cite{choi-etal-2018-ultra} represents words using the GloVe embedding \cite{pennington-etal-2014-glove} and captures semantic information of sentences, entities as well as labels with a bi-LSTM and a character-level CNN. It also learns a type label embedding matrix to operate inner product with the context and mention representation for classification. 
    \item \textbf{LabelGCN} \cite{xiong-etal-2019-imposing} improves UFET-biLSTM by stacking a GCN layer on the top to capture the latent label dependency. 
    \item \textbf{LDET} \cite{onoe-durrett-2019-learning} applies ELMo embeddings \cite{Peters:2018} for word representation and adopts LSTM as its sentence and mention encoders. Similar to UFET-biLSTM, it learns a matrix to compute inner product with each input representation for classification. Besides, LDET also trains a filter and relabeler to fix the label inconsistency in the distant supervision training data. 
    \item \textbf{BOX4Types} \cite{Onoe2021ModelingFE} introduces box embeddings to handle the type dependency problems. It uses BERT-large-uncased \cite{devlin-etal-2019-bert} as the backbone and projects the hidden classification vector to a hyper-rectangular (box) space. Each type from the label space is also represented as a box and the classification is fulfilled by computing the intersection of the input text and type boxes. 
    \item \textbf{LRN} \cite{liu2021finegrained} encodes the context and entity with BERT-base-uncased. Then two LSTM-based auto-regression network captures the context-label relation and the label-label relation via attention mechanisms respectively in order to generate labels. They simultaneously construct bipartite graphs for sentence tokens, entities and generated labels to do relation reasoning and predict more labels.
    \item \textbf{MLMET} \cite{Dai2021UltraFineET}, the prior SOTA system, first generates additional distant supervision data by BERT Masked Language Model, then stacks a linear layer on BERT to learn the classifier on the union label space.
    
    % . Their typing model uses BERT-base-cased\footnote{According to their paper, BERT-base-cased outperforms BERT-large and RoBERTa models} as the encoder. The input sentence is formatted as ``[CLS token] sentence [SEP token] entity mention [SEP token]'' and the hidden vector of [CLS token] goes through a linear classification layer to generate the probability of each type.
    % \item \textbf{RoBERTa} \cite{liu2019roberta} is added for studying the effectiveness of introducing NLI to this task. Here we adopt the same typing model of MLMET and include both of distant supervision data and human annotated data to achieve its best performance. 
\end{itemize} 

% \subsubsection{Results}\label{ssec:ultraRes} 
\begin{table}[t]
\small
\centering
\setlength{\tabcolsep}{4pt}
% \begin{tabular}{@{}llll@{}}
\begin{tabular}{ll|ccc}
% {p{0.26\linewidth}p{0.08\linewidth}p{0.08\linewidth}p{0.08\linewidth}}

\toprule
\multicolumn{2}{c|}{Model}  & \multicolumn{1}{c}{P}  & \multicolumn{1}{c}{R} & \multicolumn{1}{c}{F1}     \\ \midrule
\multicolumn{2}{l|}{UFET-biLSTM      \cite{choi-etal-2018-ultra} }                               & 48.1     & 23.2    & 31.3    \\
\multicolumn{2}{l|}{LabelGCN      \cite{xiong-etal-2019-imposing} }                           & 50.3     & 29.2    & 36.9    \\
\multicolumn{2}{l|}{LDET       \cite{onoe-durrett-2019-learning} }                              & 51.5     & 33.0    & 40.1    \\
\multicolumn{2}{l|}{Box4Types        \cite{Onoe2021ModelingFE}  }                       & 52.8     & 38.8    & 44.8    \\
\multicolumn{2}{l|}{LRN            \cite{liu2021finegrained}  }                        & \textbf{54.5}     & 38.9    & 45.4    \\
\multicolumn{2}{l|}{MLMET          \cite{Dai2021UltraFineET}  }                         & 53.6     & 45.3    & 49.1    \\ \midrule
\multirow{6}{*}{\modelname} 
&   NLI         & 1.5      & 7.1     & 2.5     \\
&\revision{L} &\revision{48.7} &\revision{45.8} &\revision{47.2}\\
&D+L       &27.5       &56.4      &37.0      \\
& NLI+D+L        & 45.4     & 49.9    & 47.4    \\
& NLI+L        & 52.4     & \textbf{48.9}    & \textbf{50.6}\\
& \enspace -- w/o label dependency & 53.3 & 46.6 & 49.7\\\bottomrule
\end{tabular}
%\vspace{-0.5em}
\caption{Results on the ultra-fine entity typing task. \modelname series are equipped with the Taxonomic Statement template. ``w/o label dependency'' is applied to the ``NLI+L'' setting. \secondrevision{The F1 result by \modelnamets\textsubscript{NLI+L} is statistically significant (p-value < 0.01 in t-test) in comparison with the best baseline result by \mbox{MLMET}.}}
\label{tab:res}
%\vspace{-1em}
\end{table}
\begin{table*}[t]
\centering
% \footnotesize
\small
\begin{tabular}{ll|cccc}

\toprule
                 &  & \multicolumn{2}{c}{OntoNotes} & \multicolumn{2}{c}{FIGER} \\ \cmidrule(l){3-4}\cmidrule(l){5-6} 
\multicolumn{2}{c|}{Model }             & {\footnotesize macro-F1}         & {\footnotesize micro-F1 }        & {\footnotesize macro-F1}       & {\footnotesize micro-F1}       \\ \midrule
\multicolumn{2}{l|}{Hierarchy-Typing \cite{chen-etal-2020-hierarchical}} & 73.0            & 68.1          & 83.0        & 79.8        \\
\multicolumn{2}{l|}{Box4Types  \cite{onoe-durrett-2020-interpretable} }       & 77.3          & 70.9          & 79.4        & 75.0        \\
\multicolumn{2}{l|}{DSAM    \cite{9305269}   }        & 83.1          & 78.2          & 83.3        & 81.5        \\
\multicolumn{2}{l|}{SEPREM   \cite{xu-etal-2021-syntax}}  &  --             &  --             & 86.1        & 82.1        \\
\multicolumn{2}{l|}{MLMET   \cite{Dai2021UltraFineET}    }        & 85.4          & 80.4          &  --           &  --           \\ \midrule
\multirow{2}{*}{\modelname} & pre-trained on NLI+UFET    & \textbf{86.6}  & \textbf{81.4}  & 80.1  & 74.7        \\
& NLI+task-specific training         & 86.4          & 80.9          & \textbf{86.7}     &\textbf{83.3}\\\bottomrule       
\end{tabular}
%\vspace{-0.5em}
\caption{Results for fine-grained entity typing. \revision{All \modelname model \secondrevision{ results are statistically significant (p-value < 0.05 in t-test) in comparison with} the best baseline results by MLMET on OntoNotes and by SEPREM on FIGER.} }
\label{tab:fineRes}
\end{table*}

\begin{table*}[ht]
\centering
\def\arraystretch{1.5}
\scriptsize
% p{0.35\linewidth}
\begin{tabular}{p{0.12\linewidth}|p{0.59\linewidth}|p{0.2\linewidth}}
\hline
Data Source   & \multicolumn{1}{c|}{Sentence}  & \multicolumn{1}{c}{Labels}                 \\ \hline
       \multirow{3}{*}{Entity Linking}       & (a) From 1928-1929 , he enrolled in graduate coursework at Yale University in New Haven , \textbf{Connecticut}.    & \textbf{location}, \textcolor{gray}{author}, \textbf{province}, \textcolor{gray}{cemetery}, \textcolor{gray}{person} \\ \cline{2-3} 

% & (b) NASA loses one of its probes, the \textbf{Mars Climate Orbiter}.                                                              & \textbf{spacecraft}, \textcolor{gray}{event}                   \\ \cline{2-3} 
            %   & (c) Berg is a municipality in the district of Rhein-Lahn , in Rhineland-Palatinate , in western \textbf{Germany}.
            %   & \textbf{country},  \textbf{location},  \textcolor{gray}{artist}, \textcolor{gray}{person} \\ \cline{2-3}
              & (b) Once Upon Andalasia is a video game based on the \textbf{film} of the same name.  & \textbf{art},  \textbf{film}\\ 
              \hline
              & (c) You can also use them in casseroles and they can be grated and fried if you want to make \textbf{hash browns}.
              & \textcolor{gray}{brown}                                        \\ \cline{2-3} 
Head Word     
% & (f) We have been hesitant at making the comparison until more evidence was widely circulated in the \textbf{mainstream media} . & \textbf{media}, \textcolor{gray}{mainstream}                                      \\ \cline{2-3} 
              & (d) He has written \textbf{a number of short stories} in different fictional worlds, including Dragonlance, Forgotten Realms, Ravenloft and Thieves' World. &\textcolor{gray}{number}     \\ \cline{2-3} 
              & (e) Despite obvious parallels and relationships , video art is not \textbf{film}.  &\textbf{film} \\ \hline
\end{tabular}
%\vspace{-0.5em}
\caption{\label{tab:dataSource} Examples of two sources of distant supervision data (one from entity linking, the other from head word extraction). In the right ``Labels'' column, \textbf{correct types} are boldfaced while \textcolor{gray}{incorrect ones} are in grey.}
%\vspace{-1em}
\end{table*}

\stitle{Results.} \Cref{tab:res} compares \modelname with baselines, in which \modelname adopts the taxonomic statement template (i.e. ``\textit{[ENTITY]} is a \textit{[LABEL]}'').  

Overall, \modelnamets\textsubscript{NLI+L} demonstrates SOTA performance over other baselines, outperforming the prior top system MLMET \cite{Dai2021UltraFineET} with 1.5\% absolute improvement on F1. Recall that MLMET built a multi-way classifier on the its newly collected distant supervision data and the human-labeled data, our \modelname optimizes a textual entailment scheme on the entailment data (i.e., MNLI) and the human-labeled entity typing data. This comparison verifies the effectiveness of using the entailment scheme and the indirect supervision from NLI.

% The results by \modelname series improve RoBERTa-large baseline method by at least 3.4\% absolute F1, showing that incorporating the inference ability of NLI has better performance than fine-tuning a classification task on pre-trained language model. 

The bottom block in \Cref{tab:res} further explores the best combination of available training data. First, training on MNLI (i.e., \modelnamets\textsubscript{NLI}) alone does not provide promising results. This could be due to that the MNLI does not generalize well to this UFET task. \revision{\modelnamets\textsubscript{L} removes the supervision from NLI as compared to \modelnamets\textsubscript{NLI+L}, causing a noticeable performance drop. In addition, the comparison} between \modelnamets\textsubscript{NLI+L} and \modelnamets\textsubscript{D+L} illustrates that the MNLI data, as an out-of-domain resource, even provides more beneficial supervision than the distant annotations. To our knowledge, this is already the first work that shows rather than relying on gathering distant supervision data in the (entity-mentioning context, type) style, it is possible to find more effective supervision from other tasks (e.g., from entailment data) to boost the performance. However, when we incorporate the distant supervision data (D) into \modelnamets\textsubscript{NLI+L}, the new system \modelnamets\textsubscript{NLI+D+L} performs worse. We present more detailed analyses in \Cref{sec:analysis}. 

In addition, we also investigate the contribution of label dependency modeling by removing it from \modelnamets\textsubscript{NLI+L}. 
% Apparently, incorporating label dependency helps improve the recall with a large margin (from 46.6 to 48.9) despite a minor drop for the precision, leading to notable overall improvement in F1.
\revision{As results shown in \Cref{tab:res}, incorporating label dependency helps improve the recall with a large margin (from 46.6 to 48.9) despite a minor drop for the precision, leading to notable overall improvement in F1.}

% \wenpeng{ Discuss it here: To understand how dependency modeling influence the typing results, we investigate the performance breakdown  of \modelnamets\textsubscript{w/o aug} of different label specificity on the test set. In \Cref{tab:breakdown}, label dependency modeling mostly improves \modelnamets\textsubscript{w/o aug} on the fine-grained types by 2.6\% in F1. This is because fine-grained labels, as the most infrequent labels, only takes up 12.1\% of the label occurrences in the human annotated training set. During the dependency modeling phase, type descriptions of ultra-fine grained labels, which in total take up 70.1\% of the training labels, therefore provide indirect supervision to these infrequent labels. General labels also benefit from this mechanism with 2.6\% absolute improvement.} 
% For the different training data settings of \modelname, the performance of \modelnamets\textsubscript{NLI} indicates that simply training the language model on MNLI cannot capture the sentences and type descriptions well. Training with typing data endows the model with ability to distinguish between correct and false type descriptions. \bangzheng{\modelnamets\textsubscript{D+L}}

\subsection{Fine-grained Entity Typing}\label{ssec:fine} 
In addition to UFET, we are also interested in (i) the effectiveness of our \modelname to entity typing tasks with much fewer types, and (ii) if our learned \modelname model from the ultra-fine task can be used for inference on other entity typing tasks, \textit{which often has unseen types}, even without further tuning. To the end, we evaluate \modelname on OntoNotes \cite{gillick2014context} and FIGER \cite{ling2012fine}, two popular fine-grained entity typing benchmarks.

\textbf{OntoNotes} contains 3.4M automatically labeled entity mentions for training and 11k manually annotated instances that are split into 8k for dev set and 2k for test set. Its label space consists of 88 types and one more $other$ type. In inference, \modelname outputs $other$ if none of the 88 types is scored over the threshold described in \Cref{ssec:inf}.
% In the label space of OntoNotes, there is a frequent label \textit{other} that produces meaningless type descriptions for \modelname. Though its descendant types are meaningful words, \textit{other} sometimes appears alone to label the entity mentions that is difficult to be described as the other 88 words. For example, entity \textit{Friday} is therefore labeled as a single \textit{other}. To alleviate this problem, we modified \modelname to produce a single \textit{other} if no type descriptions is scored over the threshold described in \Cref{ssec:inf}.
\textbf{FIGER} contains 2M data samples labeled with 113 types. The dev set and test set include 1,000 and 562 samples respectively. Within its label space, 82 types have a dependency relation with their ancestor or descendant types while the other 30 types are uncategorized free-form words. 

% \stitle{Baselines}
%  We compare \modelname with 5 following approaches: Hierarchy-Typing \cite{chen-etal-2020-hierarchical}, Box4Types \cite{onoe-durrett-2020-interpretable}, DSAM\cite{9305269}, MLMET\cite{Dai2021UltraFineET} and K-ADAPTER \cite{DBLP:journals/corr/abs-2002-01808}. 
%  Hierarchy-Typing designed a hierarchy-aware ranking loss that encourages scoring descendant types over ancestor types. It represents the input tokens with ELMO \cite{Peters:2018} embeddings and generates the context and mention embeddings with attention mechanism. 
%  DSAM utilizes a diversified semantic attention mechanism to capture both context-level and mention-level information. 
%  K-ADAPTER introduces a plug-in neural adaptor that infuse factual and linguistic knowledge into the pre-trained language model. 
%  All of these three models deploy a classification layer as the decoder. 

% For this benchmark, we compare \modelname with the following methods: ERNIE \cite{zhang-etal-2019-ernie}, WKLM \cite{Xiong2020pre-trained}, DSAM \cite{9305269} and K-ADAPTER \cite{DBLP:journals/corr/abs-2002-01808}. ERNIE takes entity embeddings pre-trained from knowledge graphs as inputs to the Transformer so as to capture the entity relations. Similarly, K-ADAPTER introduces a plug-in neural adaptor that infuse factual and linguistic knowledge into the pre-trained language model. WKLM replaces the entity mention in the sentence with different words of same type for knowledge learning.

% \input{tables/ontoResult}
% \input{tables/figerResult}
% \bangzheng{Ongoing Results}
\stitle{Results.} \Cref{tab:fineRes} reports baseline results as well as results of two variants of \modelname: one is pre-trained on UFET and directly transfer to predict on the two target benchmarks, the other conducts task-specific training on the target benchmark after pre-training on MNLI. 
The task-specific training variant outperforms respective prior SOTA on both benchmarks (OntoNotes: 86.4 vs. 85.4 in macro-F1, 80.9 vs. 80.4 in micro-F1; FIGER: 86.7 vs. 84.9 in macro-F1, 83.3 vs. 81.5 in micro-F1). 

\revision{An interesting advantage of \modelname lies in its \textbf{transferability} across benchmarks.  \Cref{tab:fineRes} demonstrates that our \modelname (pre-trained on UFET) offers competitive performance on both OntoNotes and FIGER even with only zero-shot transfer (it even exceeds the ``task-specific training'' version on OntoNotes).\footnote{\modelname pre-trained on UFET performs worse on FIGER than \modelname with task-specific training. The main reason could be that a larger portion of FIGER test data comes with an entity of proper noun to be labeled with more compositional types, such as \textit{government agency, athlete, sports facility}, which has appeared much less on UFET.} Although there are disjoint type labels between these two datasets and UFET, there exist manually-crafted mappings from UFET labels to them}
\secondrevision{(e.g. ``musician'' to ``/person/artist/music'').}
\revision{In this way, traditional multi-way classifiers still work across the datasets after type mapping though we do not prefer human-involvement in real-world applications. To further test the transferability of \modelname, a more challenging experimental setting for zero-shot type prediction is conducted and analyzed in \Cref{sec:analysis}.}

\begin{table*}[t]
\centering
\small
\setlength{\tabcolsep}{5pt}
\begin{tabular}{@{}l|lllllllll@{}}
\toprule
& \multicolumn{3}{c}{\modelnamets\textsubscript{NLI+L}}   & \multicolumn{3}{c}{\modelnamets\textsubscript{NLI+D+L}}   & \multicolumn{3}{c}{\modelnamets\textsubscript{D+L}}  \\
\cmidrule(lr){2-4} \cmidrule(lr){5-7} \cmidrule(lr){8-10}
Templates & P       & R      & F1     & P       & R       & F1     & P      & R      & F1\\ 
\midrule
Taxonomic Statement   & 52.4	& 48.9	&\textbf{50.6} &45.4 &49.9 &\textbf{47.4}   &  27.5	&56.4	&\textbf{37.0}\\ 
Contextual Explanation  &50.8	&49.2	&50.2	&45.3 &48.5	&46.8	&26.9	&55.4	&36.2\\
Label Substitution &47.4	&49.3	&48.3	&42.5 &50.7	&46.2	&24.8	&59.3	&35.0\\
\bottomrule
\end{tabular}
%\vspace{-0.5em}
\caption{\label{tab:breakdown} Behavior of different type description templates under three training settings. }
%\vspace{-0.5em}
\end{table*}

\subsection{Analysis}\label{sec:analysis} 

Through the following analyses, we try to answer \revision{following} questions: (i) Why did not the distant supervision data help (as \Cref{tab:res} indicates)? (ii) How effective is each type description template (\Cref{tab:typeDes})? (iii) With the NLI-style formulation and the indirect supervision, does \modelname generalize better for zero-shot and few-shot prediction? \revision{Is trained \modelname transferable to new benchmarks with unseen types?}  (iv) On which entity types does our model perform better, and which ones remain challenging? \revision{(vi) How efficient is \modelname?}

% In this section, we provide analyses on different training settings and model performances including the effect of training data on \modelname, the few-shot prediction ability of \modelname, the effectiveness of dependency modeling as well as the benefits of leveraging language inference into typing task. Case Study are also provided for explanation. 

\stitle{Distant Supervision Data.}
As \Cref{tab:res} indicates, adding distant supervision data in \modelnamets\textsubscript{NLI+D+L} even leads to a drop of 3.2\% absolute score in F1 from \modelnamets\textsubscript{NLI+L}. This should be due to the fact that the distant supervision data (D) are overall noisy \cite{onoe-durrett-2019-learning}. \Cref{tab:dataSource} lists some \textit{frequent and typical} problems that exist in D based on  entity linking and head-word extraction. In general, they will lead to two problems. 

On the one hand, a large number of false positive types are introduced. Considering the example (a) in \Cref{tab:dataSource}, the state \textit{Connecticut} is labeled as \textit{author}, \textit{cemetery} and \textit{person}. For the example (c), \textit{hash brown} is labeled as \textit{brown}, turning the concept of food into color. Additionally, the head-word method is short in capturing the semantics. In the example (d), \textit{number} is falsely extracted as the type for \textit{a number of short stories} because of the preposition ``of''. 

On the other, such distant supervision may not comprehensively recall positive types. For instance, examples (b) and (e) are both about the entity ``film'' where the recalled types are correct. However, in the human annotated data, entity ``film'' may also be labeled as (``film'', ``art'',  ``movie'', ``show'', ``entertainment'', ``creation''). In this situation, those missed positive types (i.e., ``movie'', ``show'', ``entertainment'' and ``creation'') will be  selected by the negative sampling process of \modelname and therefore negatively influence the performance. The comparison between  \modelnamets\textsubscript{NLI+L} and \modelnamets\textsubscript{D+L} can further justify the superiority of the indirect supervision from NLI over that from the distant supervision data.

\stitle{Type Description Templates.} 
\Cref{tab:breakdown} reveals how template choices affect the typing performance. It is obvious that taxonomic statement outperforms the other two under all of the three training settings. The contextual explanation template yields close, while worse results but the label substitution leads to more noticeable F1 drop. \revision{This may result from the absence of entity mention in hypothesis by label substitution. For instance, in \textit{``\underline{Soft eye shields} are placed on the babies to protect their eyes.''}, \modelname with label substitution generates related but incorrect type labels such as \textit{treatment}, \textit{attention} or \textit{tissue}. }
% The main reason for this phenomenon comes from the absence of entity mention in the type descriptions by label substitution template. In this case, incorrect candidate type words can produce plausible hypothesis for NLI and yield wrong typing result. For example, in ``\underline{\textbf{Soft Eye shields}}''\modelname with label substitution template predicts erroneous labels \textit{treatment}, \textit{tissue} and \textit{safety glasses}. With no constraints from \textbf{soft eye shields}, type descriptions of correct labels are more difficult for \modelname to distinguish from other ``covering-like'' nouns. 

% The noisy labels in entity linking data mainly consist of \textit{unrelated types}. For example, in the sample (a), entity \textit{Conneticut} is here labeled as \textit{author}, \textit{cemetery} and \textit{person}. In contrast, the head-word extraction can result in \textit{incorrect and incomplete types}. For the example (c), entity \textit{hash brown} is labeled as \textit{brown}, turning the concept of food into color. Another typical problem is that, the head-word based  method is short in capturing the semantics. In the example (d), \textit{number} is extracted as the type for \textit{a number of short stories} because of the preposition ``of''. 

\begin{figure}[t]
\centering
\includegraphics[width=0.9\columnwidth]{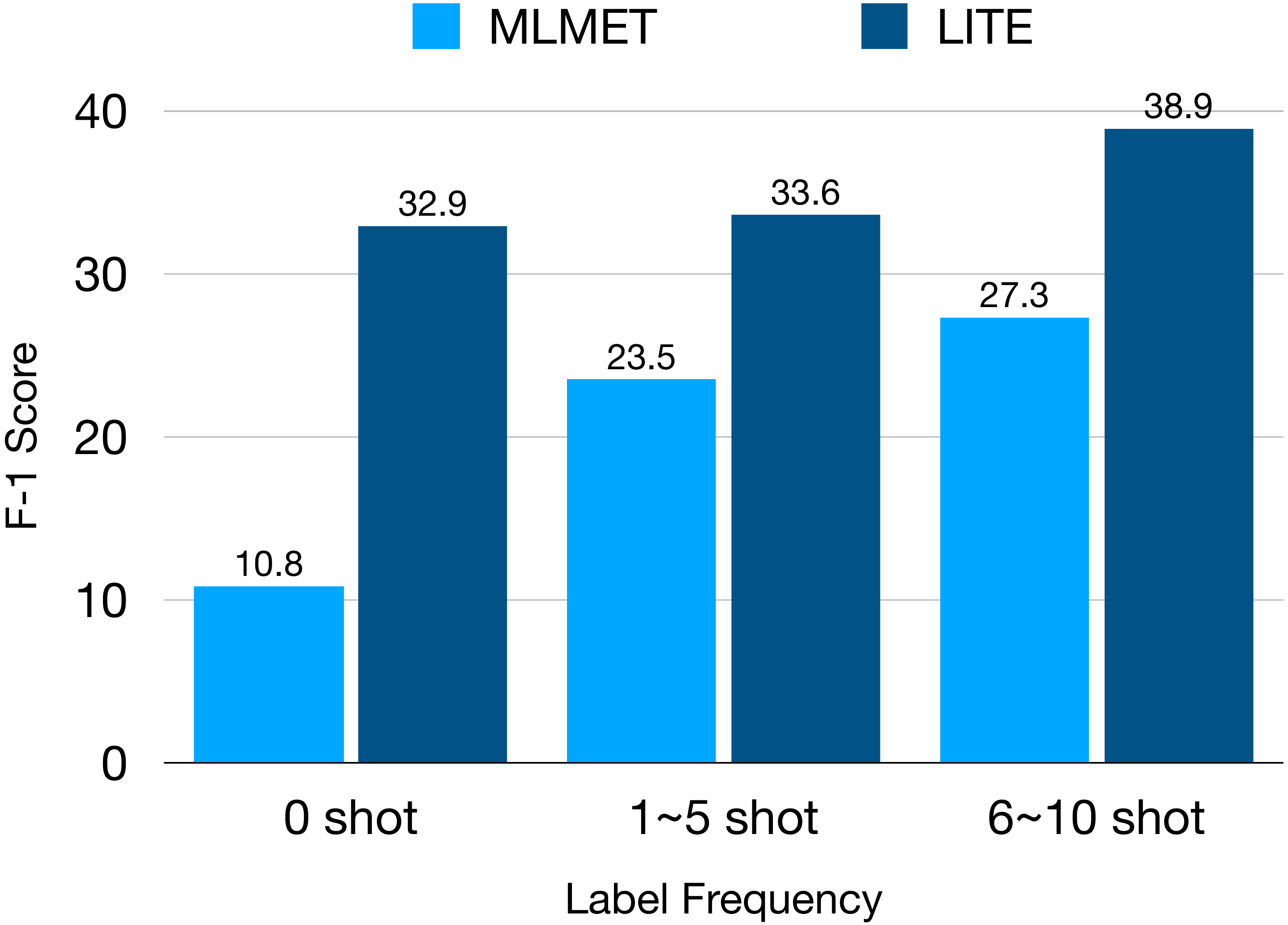}
%\vspace{-1em}
\caption{\revision{Performance comparison of our system \modelname and the prior SOTA system, MLMET, on the filtered version of UFET for zero-shot and few-shot typing. The zero-shot labels correspond to the 40\% test set type labels that are unseen in training. We also report the performance on other few-shot type labels.}}
\label{fig:fewShot}
%\vspace{-1em}
\end{figure}
% \usepackage{booktabs}
\begin{table*}[t]
\centering
\scriptsize
\begin{tabular}{c|p{0.3\linewidth}|p{0.2\linewidth}|p{0.37\linewidth}}
\toprule
& Input       & True Labels       & Prediction\\ \midrule\midrule
% (a) \textbf{Soft eye shields} are placed on the baby to protect their eyes from damage that may lead to retinopathy due to the bili lights.                               & device, shield, object, guard, protection, defender, shade, object                                          & \begin{tabular}[c]{p{1\linewidth}}Taxonomic Statement: device, shield, object, protection, covering\\ Contextual Explanation: shield, object, protection, safeguard, defense\\ Label Substitution: treatment, tissue,  safety glasses, item, device, shield, object\end{tabular} \\ \midrule\midrule
\multirow{13}{*}{\rotatebox{90}{\modelname exceeds MLMET}}& (a) The University of California   communications major gave \textbf{\underline{her}} mother a  fitting present, surprising herself by  winning the 50-meter backstroke gold medal.  & athlete, person,\newline swimmer*, contestant, \newline scholar, child                                                       & 
\modelname: \newline
 \textcolor{yaleblue}{athlete}, \textcolor{yaleblue}{person}, \textcolor{yaleblue}{swimmer*}, female, student,  winner  \newline 
MLMET: \newline
\textcolor{yaleblue}{athlete}, \textcolor{yaleblue}{person}, \textcolor{yaleblue}{child}, adult, female, mother, woman \\ \cmidrule(lr){2-4}

& (b) The apology is being viewed as a watershed in Australia , with major television networks airing \textbf{\underline{it}} live and crowd gathering around huge screens in the city. &  event, apology*,\newline   plea, regret     
& \modelname: \newline \textcolor{yaleblue}{event}, \textcolor{yaleblue}{apology*}, ceremony, happening, concept \newline MLMET: \newline \textcolor{yaleblue}{event}, message \\ \cmidrule(lr){2-4}

& (c) A drawing table is also sometimes called a mechanical desk because , for several centuries , most \textbf{\underline{mechanical desks}} were drawing tables.    & object, desk, \newline furniture*, board, \newline desk, table       & \modelname: \newline \textcolor{yaleblue}{object}, \textcolor{yaleblue}{desk}, \textcolor{yaleblue}{furniture*}\newline  MLMET: \newline \textcolor{yaleblue}{object}, \textcolor{yaleblue}{desk}, computer \\ \midrule\midrule

\multirow{9}{*}{\rotatebox{90}{MLMET exceeds \modelname}} & (d) He attended the University of Virginia , where he played \textbf{\underline{basketball and baseball}} ; his brother Bill also played baseball for the University.                                &  basketball*, baseball, fun,\newline action, activity,  contact sport, game, sport, athletics, ball game, ball, event & \modelname:\newline  \textcolor{yaleblue}{activity}, \textcolor{yaleblue}{game}, \textcolor{yaleblue}{sport}, \textcolor{yaleblue}{event}, \textcolor{yaleblue}{ball game}, \textcolor{yaleblue}{ball}, \textcolor{yaleblue}{athletics}\newline  MLMET:\newline \textcolor{yaleblue}{activity}, \textcolor{yaleblue}{game}, \textcolor{yaleblue}{sport},  \textcolor{yaleblue}{event},  \textcolor{yaleblue}{basketball*} \\ \cmidrule(lr){2-4}

& (e) The \textbf{\underline{manner in which it was confirmed}} however smacked of an acrimonious end to the relationship between club and player with Chelsea.                                         & manner*, way, concept, style, method                                                                        & \modelname:\newline  event\newline MLMET: \newline \textcolor{yaleblue}{manner*}, event              \\ \bottomrule                                                                                                                                                                                         
\end{tabular}
%\vspace{-0.5em}
\caption{Case Study of labels on which \modelname improves MLMET or MLMET outperforms \modelname. \textcolor{yaleblue}{Correct predictions} are in blue and * indicates the representative label words for the discussed pattern. }
\label{tab:caseStudy}
\end{table*}

\begin{table*}[t]
\centering
\small
\setlength{\tabcolsep}{5pt}
\begin{tabular}{@{}l|lllllllll@{}}
\toprule
& \multicolumn{3}{c}{Named Entity}   & \multicolumn{3}{c}{Pronoun}   & \multicolumn{3}{c}{Nominal}  \\
\cmidrule(lr){2-4} \cmidrule(lr){5-7} \cmidrule(lr){8-10}
  & P       & R      & F1     & P       & R       & F1     & P      & R      & F1\\ 
\midrule
\modelnamets\textsubscript{NLI+L}   & \textbf{58.6}	&\textbf{55.5}	&\textbf{57.0} &51.2 &\textbf{57.5} &\textbf{54.2}   &  45.3 &\textbf{47.1} &\textbf{46.2}\\ 
MLMET &58.3 &54.4 &56.3 &\textbf{57.2} &50.0 &53.4 &\textbf{49.5} &38.9 &43.5\\
% Contextual Explanation  &50.8	&49.2	&50.2	&45.3 &48.5	&46.8	&26.9	&55.4	&36.2\\
% Label Substitution &47.4	&49.3	&48.3	&42.5 &50.7	&46.2	&24.8	&59.3	&35.0\\
\bottomrule
\end{tabular}
%\vspace{-0.5em}
\caption{\label{tab:pronoun} \revision{Performance comparison of \modelname and prior SOTA, MLMET, on named entity, pronoun and nominal entities respectively.}}
%\vspace{-0.5em}
\end{table*}

% \begin{figure}[t]
% \centering
% \includegraphics[width=0.9\columnwidth]{figures/split.pdf}
% %\vspace{-1em}
% \caption{\revision{Performance comparison of our system \modelname and the prior SOTA MLMET on the filtered version of UFET where 40\% test set type labels are unseen in training.}}
% \label{fig:split}
% %\vspace{-1em}
% \end{figure}
% \input{tables/caseStudy}

% Another essential factor for the performance drop in \modelnamets\textsubscript{NLI+D+L} is the incomplete typing labels provided by the distant supervision data. This is common for both of the entity linking and head word extraction data. For instance, example (d) and (g) are about the entity ``film'' where the label words are correct. However, in the human annotated data, entity ``film'' will probably be labeled as (``film'', ``art'',  ``movie'', ``show'', ``entertainment'', ``creation''). In this situation, the unlabeled (``movie'', ``show'', ``entertainment'', ``creation'') can be selected by the negative sampling process of \modelname and therefore negatively influence the performance. 

\stitle{Few- \& Zero-shot Prediction.} \revision{In \Cref{ssec:fine}, we discussed about transferring \modelname trained on UFET to other fine-grained entity typing benchmarks. Nevertheless, since UFET labels are still inclusive of them with mapping, we conducted further experiment in which portions of UFET training labels are randomly filtered out so that 40\% of the testing labels are unseen in training. We then investigated the \modelnamets\textsubscript{NLI+L} performance on test types which have zero or a few labeled examples in the training set.} 
% There is a non-overlapping set of types between the human annotated UFET training set and test set. Given this fact, we can investigate the \modelnamets\textsubscript{NLI+L} performance on test types which have zero or a few labeled examples in the training set. 
\Cref{fig:fewShot} shows the results of \modelnamets\textsubscript{NLI+L} and the strongest baseline, MLMET. 
% Recall that is trained with abundant training resources of LM-augmented data, distant supervision data and human annotated data, whereas 52 out of 549 zero-shot labels are still missing in their training data. 
\revision{Note that while the held-out set of type labels are completely unseen to \modelname, the full type vocabulary is however provided for MLMET during its LM-based data augmentation process in this experiment.} 

As shown in the results,
it is as expected that the performance on more frequent labels are better than on rare labels. \modelnamets\textsubscript{NLI+L} outperforms MLMET on all the listed frequency of labels which reveals the strong low-shot prediction performance of our model. Particularly, on the extremely challenging zero-shot labels, \modelnamets\textsubscript{NLI+L} drastically exceeds MLMET by 32.9\% vs. 10.8\% in F1. %Given these facts can we conclude that the template-based type description is promising in encoding rare types in context and it  make effective use of the inference ability of NLI to recognize those rare types. 
Hence, it is demonstrated that the NLI-based entity typing succeeds in more reliably representing and inferring rare and unseen entity types. 

\revision{The main difference between the NLI framework and multi-way classifiers is NLI makes use of the semantics of input text as well as the label text; conventional classifiers, however, only model the semantics of input text. Encoding the semantics of labels’ side is particularly beneficial when the type set is super large and many types lack training data. When some test labels are filtered out in the training process, \modelname still performs well with its inference manner but classifiers (like MLMET) fail to recognize the semantics of unseen labels merely with their features. In this way, \modelname maintains high performance when transfers across benchmarks with disjoint type vocabularies.}

\stitle{Case Study.} 
% We calculate the F1 scores for each label in the UFET test set and compare the results with MLMET to demonstrate how inference ability of NLI as well as type descriptions improves the prior best method. Besides, we also investigate on labels that \modelname perform worse than MLMET and the statistics are presented in \Cref{figure:diffF1}. The x-axis indicates the difference of F1 scores  (i.e. To what magnitude does \modelname improves MLMET and vice versa.)\bangzheng{dont know if this works or just delete these...} 
We randomly sampled 100 labels on which \modelname improves MLMET by at least 50\% in F1 and here are the recognized typical patterns: 
%  100 labels on which MLMET outperforms \modelname00 labels conclude the three typical circumstances that \modelname drastically outperforms MLMET:
\begin{itemize}[leftmargin=1em]
\setlength\itemsep{0em}
    \item \textbf{Contextual inference (28\%)}: In case (a) of \Cref{tab:caseStudy}, considering the information ``wining the 50-meter backstroke gold medal'', \modelname successfully types \underline{\textbf{her}} with \textit{swimmer} in addition to \textit{athlete} that is given by MLMET.
    \item \textbf{Coreference (20\%)}: In case (b), \modelname correctly refers the pronoun entity \textbf{it} to ``apology'' but MLMET merely captures local information ``tv network airing'' to obtain the label words \textit{event, message}.
    \item \textbf{Hypernym (19\%)}: %\bangzheng{if this name is proper?} 
    In the case (c), even if there is no mention of \textit{furniture} in the text, \modelname gives a high confidence score to this type that is a hypernym of \textbf{mechanical desks}. Nevertheless, MLMET only get trivial answers such as \textit{desk, object}.
\end{itemize}
On the other hand, we also sampled 100 labels on which MLMET performs better and it can be concluded that \modelname falls short mainly in following scenarios:
\begin{itemize}[leftmargin=1em]
\setlength\itemsep{0em}
    \item \textbf{Multiple nominal words (30\%)}: In the sample (d) of \Cref{tab:caseStudy}, due to ambiguous meaning of the type hypothesis ``\underline{\textbf{basketball and baseball}} is a \textit{basketball}'', \modelname fails to predict the groundtruth label \textit{basketball}.
    \item \textbf{Clause (28\%)} Instance (e) illustrates a common situation when clauses are included in the entity mention, where the effectiveness of type descriptions is harmed. The clausal information distracts \modelname from focusing on the key part of the entity.
\end{itemize}
% We present a case study with three representative cases for \modelnamets\textsubscript{Direct} with three templates (\Cref{tab:caseStudy}). In the first example, Taxonomic Statement and Contextual Explanation based models correctly predict part of the human annotation and labeled \textit{soft eye shields} with \textit{covering} and \textit{defense} respectively, which is reasonable from human's recognition. However, Label Substitution presents trivial label \textit{item} and wrong label \textit{tissue}. For the second example, three template based models correctly types the entity with the words from human annotated labels. Their false positive labels such as \textit{play}, \textit{film} and \textit{entertainment} are compatible with the given context. Nevertheless, none of them can predict a label other than \textit{person} in the third sentence. From human's perspective, it is natural to assume \textit{he} is an \textit{economist} or an \textit{administrator} while \modelname failed to capture such information.  
\revision{\stitle{Prediction on Different Categories of Entity Mentions.}}
\revision{We also investigated the prediction of \modelname on three different categories of entity mentions from the UFET test data: named entities, pronouns and nominals. For each category of mentions, we randomly sample 100 instances and the performance comparison against MLMET is reported in \Cref{tab:pronoun}.} 

\revision{According to the results, \modelname consistently outperforms MLMET on all three categories of entities and the improvement on nominal phrases (46.2\% vs 43.5\% in F1) is most significant. 
This partly aligns with the capability of making inference based on noun hypernyms, as being discussed in Case Study.
Meanwhile, typing on nominals seeks to be more challenging than on the other two categories of entities, which, from our observation, is mainly due to two reasons. 
First, Nominal phrases with multiple words are more difficult to capture by the language model in general. 
Second, nominals are sometimes less concrete than pronouns and named entities, hence \modelname also generates more abstract type labels. For example, \modelname has labeled \textbf{\underline{the drink}} in an instance as \textbf{substance}, which is too abstract and is not recognized by human annotators.} 

\revision{\stitle{Time Efficiency.}} 
\revision{In general, \modelname has much less training cost, of around 40 hours, than the previous strongest (data-augmentation-based) model MLMET, which requires over 180 hours, on the UFET task.\footnote{\secondrevision{All the time estimations are given by experiments on a commodity server with a TITAN RTX. Training and evaluation batch sizes are maximized to 16 or 128 for \modelname and MLMET respectively.}} During the inference step, it takes about 35 seconds per new sentence for our model to do inference with a fixed type vocabulary of over 10,000 different labels while a common multi-way classifier merely requires around 0.2 seconds. In fact, such a big difference in inference cost results from encoding longer texts and multiple encoding calculation for the same text. It can be accelerated by modifying the encoding model structure which will be discussed in \Cref{sec:future}. 
However, \modelname is much more efficient on dynamic type vocabulary. It requires almost no re-calculation when new, un-mappable labels are added to an existing type set but multi-way classifiers need re-training with an extended classifier every time (e.g. over 180 hours by the previous SOTA). } 

\section{Conclusion and Future Work}
\label{sec:future}
We propose a new model \modelname that leverages indirect supervision from NLI to type entities in texts. Through template-based type hypothesis generation, \modelname formulates the entity typing task as a language inference task and meanwhile the semantically rich hypothesis remedy the data scarcity problem in the UFET benchmark. 

Besides, the learning-to-rank objective further help \modelname with generalized prediction across benchmarks with disjoint type sets. Our experimental results illustrate that \modelname promisingly offer SOTA on UFET, OntoNotes and FIGER, and yields strong performance on zero-shot and few-shot types. \modelname pretrained on UFET also yields strong transferability by outperforming SOTA baselines when directly make predictions on OntoNotes and FIGER. 

\revision{For future research, as mentioned in \Cref{sec:analysis}, we first plan to investigate ways to accelerate \modelname by utilizing a late-binding cross-encoder \cite{pang-etal-2020-fastmatch} for linear-complexity NLI, and incorporating high-dimensional indexing techniques like ball trees in inference. To be specific, the premise and hypotheses can first be encoded respectively and the resulting representations can later be used to evaluate the confidence score of premise-hypothesis representation pairs through a trained network. With little expected loss in performance, \modelname can still maintain its feature of strong transferability and zero-shot prediction.}

\revision{In addition,} we plan to extend NLI-based indirect supervision to information extraction tasks such as relation extraction and event extraction. Incorporating abstention-awareness \cite{Network_Agnostophobia} for handling unknown types is another meaningful direction. 
\revision{Besides, \citet{poliak-etal-2018-collecting} recasted diverse types of reasoning dataset including NER, relation extraction and sentiment analysis into NLI structure, which we plan to incorporate as extra indirect supervision for \modelname to further enhance the robustness of entity typing.}

\section*{Acknowledgment}

The authors appreciate the reviewers and editors for their insightful comments and suggestions. The authors would also like to thank Hongliang Dai and Yangqiu Song from the Hong Kong University of Science and Technology for sharing the resources and implementation of MLMET,
and thank Eunsol Choi from the University of Texas at Austin for sharing the full UFET distant supervision data.

This material is partly supported by the National Science Foundation of United States Grant IIS 2105329, and the DARPA MCS program under Contract No. N660011924033 with the United States Office Of Naval Research.

\bibliography{tacl2018}
\bibliographystyle{acl_natbib}

\end{document}